\documentclass[10pt, conference]{IEEEtran}
\IEEEoverridecommandlockouts
\usepackage{cite}
\usepackage{textcomp}
\usepackage{color,xcolor}
\usepackage{epsfig}
\usepackage{graphicx}

\usepackage{adjustbox}
\usepackage{array}
\usepackage{booktabs}
\usepackage{colortbl}
\usepackage{float,wrapfig}
\usepackage{hhline}
\usepackage{multirow}
\usepackage{subcaption} 

\usepackage{amsmath,amsfonts,amsthm,amssymb}
\usepackage{bm}
\usepackage{nicefrac}
\usepackage{microtype}

\usepackage{changepage}
\usepackage{extramarks}
\usepackage{fancyhdr}
\usepackage{setspace}
\usepackage{soul}
\usepackage{xspace}

\usepackage[ruled]{algorithm2e} 
\usepackage{xcolor} 
\usepackage{tikz}
\usetikzlibrary{fit,calc}
\usepackage{enumerate}
\usepackage{pifont}
\usepackage{hyperref}

\newcolumntype{L}[1]{>{\raggedright\let\newline\\\arraybackslash\hspace{0pt}}m{#1}}
\newcolumntype{C}[1]{>{\centering\let\newline\\\arraybackslash\hspace{0pt}}m{#1}}
\newcolumntype{R}[1]{>{\raggedleft\let\newline\\\arraybackslash\hspace{0pt}}m{#1}}

\newcommand{\ignorethis}[1]{}

\makeatletter
\DeclareRobustCommand\onedot{\futurelet\@let@token\@onedot}
\def\@onedot{\ifx\@let@token.\else.\null\fi\xspace}

\def\eg{\emph{e.g}\onedot}

 \def\vs{\emph{vs}\onedot}
 
\def\etal{\emph{et al}\onedot}
\makeatother

\definecolor{mydarkblue}{rgb}{0,0.08,1}
\definecolor{mydarkgreen}{rgb}{0.02,0.6,0.02}
\definecolor{mydarkred}{rgb}{0.8,0.02,0.02}
\definecolor{mydarkorange}{rgb}{0.40,0.2,0.02}
\definecolor{mypurple}{RGB}{111,0,255}
\definecolor{myred}{rgb}{1.0,0.0,0.0}
\definecolor{mygold}{rgb}{0.75,0.6,0.12}
\definecolor{mydarkgray}{rgb}{0.66, 0.66, 0.66}
\definecolor{mygray}{gray}{0.9}

\def\x{$\times$}
\def\X{$\times$\xspace}

\def\naas{NAAS\xspace}
\def\NAAS{Neural Accelerator Architecture Search\xspace}

\def\BibTeX{{\rm B\kern-.05em{\sc i\kern-.025em b}\kern-.08em
    T\kern-.1667em\lower.7ex\hbox{E}\kern-.125emX}}
\begin{document}

\title{\naas: \NAAS}

\author{
\IEEEauthorblockN{Yujun Lin\textsuperscript{*}}
\IEEEauthorblockA{\textit{MIT}}
\and
\IEEEauthorblockN{Mengtian Yang\textsuperscript{*}}
\IEEEauthorblockA{
\textit{SJTU}\\
\textcolor{mydarkblue}{\texttt{\url{https://tinyml.mit.edu}}}
}
\and
\IEEEauthorblockN{Song Han}
\IEEEauthorblockA{\textit{MIT}}
\thanks{\textsuperscript{*}Equally contributed to this work.}
}

\IEEEoverridecommandlockouts
\IEEEpubid{\makebox[\columnwidth]{978-1-6654-3274-0/21/\$31.00~\copyright2021 IEEE \hfill} \hspace{\columnsep}\makebox[\columnwidth]{ }}
\maketitle
\IEEEpubidadjcol

\begin{abstract}
Data-driven, automatic design space exploration of neural accelerator architecture is desirable for specialization and productivity.
Previous frameworks focus on sizing the numerical architectural hyper-parameters while neglect searching the PE connectivities and compiler mappings.
To tackle this challenge, we propose \NAAS (\naas) that holistically searches the neural network  architecture, accelerator architecture and compiler mapping in one optimization loop. \naas composes highly matched architectures together with efficient mapping. 
As a data-driven approach, \naas rivals the human design Eyeriss by 4.4$\times$ EDP reduction with 2.7\% accuracy improvement on ImageNet under the same computation resource, and offers 1.4$\times$ to 3.5$\times$ EDP reduction than only sizing the architectural hyper-parameters.
\end{abstract}
\section{Introduction}
Neural architecture and accelerator architecture co-design is important to enable specialization and acceleration. 
It covers three aspects: designing the neural network, designing the accelerator, and the compiler that maps the model on the accelerator. 
The design space of each dimension is listed in Table~\ref{tab:enhas-space}, with more than $10^{800}$ choices for a 50-layer neural network. Given the huge design space, data-driven approach is desirable, where new architecture design evolves as new designs and rewards are collected. 
Recent work on hardware-aware neural architecture search (NAS) and auto compiler optimization have successfully leverage machine learning algorithms to automatically explore the design space.
However, these works only focuses on off-the-shelf hardware~\cite{cai2018proxylessnas, lu2019neural, wang2019haq, cai2020once, chen2018learning, kao2020gamma}, and neglect the freedom in the hardware design space.

The interactions between the neural architecture and the accelerator architecture is illustrated in Table~\ref{tab:relationship}. 
The correlations are complicated and vary from hardware to hardware: for instance, tiled input channels should be multiples of \#rows of compute array in NVDLA, while \#rows is related to the kernel size in Eyeriss. It is important to consider all the correlations and make them fit. A tuple of perfectly matched \textit{neural architecture}, \textit{accelerator architecture}, and \textit{mapping strategy} will improve the utilization of the compute array and on-chip memory, maximizing efficiency and performance.

The potential of exploring both neural and accelerator architecture has been proven on FPGA platforms~\cite{hao2019fpga, li2020edd, zhang2019skynet, kao2020confuciux} where HLS is applied to generate FPGA accelerator. Earlier work on accelerator architecture search~\cite{yang2020co, lin2019nhas, jiang2020hardware} only search the architectural sizing while neglecting the PE connectivity (\eg, array shape and parallel dimensions) and compiler mappings, which impact the hardware efficiency.

\begin{table}[t]
  \centering
  \caption{Neural-Accelerator architecture search space.}
    \scalebox{0.75}{
    \begin{tabular}{c|l}
    \toprule
    \multirow{3}[2]{*}{Accelerator } &Compute Array Size ($\#$rows/$\#$columns) \\
          &  (Input/Weight/Output) Buffer Size \\
          &  PE Inter-connection (Dataflow) \\
    \midrule
    Compiler Mapping & Loop Order, Loop Tiling Sizes\\
    \midrule
    \multirow{2}[2]{*}{Neural Network} & \#Layers, \#Channels, Kernel Size \\
          & Block Structure, Input Size\\
    \bottomrule
    
    \end{tabular}%
    }
  \label{tab:enhas-space}%
\end{table}%
\begin{table}[t]
  \centering
  \caption{The complicated correlation between neural and accelerator design space. It differs from accelerator to accelerator: N is NVDLA and E is Eyeriss.}
  \scalebox{0.8}{
    \begin{tabular}{c|cccc}
    \toprule
 Accelerator & \multicolumn{4}{c}{Neural Architecture Design Space} \\
\cmidrule{2-5}  Parameter & Input & Output & Kernel & Feature \\
  Space  & Channels & Channels & Size  & Map Size \\
    \midrule
    Array \#rows & N   &     &  E  &  \\
    \midrule
    Array \#cols &     & N   &     & E\\
    \midrule
    IBUF Size    & N/E &     &     & N/E \\
    \midrule
    WBUF Size    & N/E & N/E & N/E &  \\
    \midrule
    OBUF Size    &     & N/E &     & N/E\\
    \bottomrule
    \end{tabular}
    }
  \label{tab:relationship}%
\end{table}%

We push beyond searching only hardware hyper-parameters and propose the \NAAS (\naas), which fully exploits the hardware design space and compiler mapping strategies at the same time. Unlike prior work~\cite{yang2020co} which formulate the hardware parameter search as a pure sizing optimization, \naas models the co-search as a two-level optimization problem, where each level is a combination of indexing, ordering and sizing optimization. To tackle such challenges, we propose an encoding method which is able to encode the \textit{non-numerical} parameters such as loop order and parallel dimension chosen as \textit{numerical} parameters for optimization. As shown in Figure~\ref{fig:naas},  the outer loop of \naas optimizes the accelerator architecture while the inner loop optimizes the compiler mapping strategies. 

Combining both spaces greatly enlarges the optimization space: within the same \#PEs and on-chip memory resources as EdgeTPU there are at least $10^{11}$ hardware candidates and $10^{17}$ mapping candidates for each layer, which composes $10^{11+50\cdot17}=10^{861}$ possible combinations in the joint search space for ResNet-50, while there are only $10^4$ hardware candidates in NASAIC's design space. To efficiently search over the large design space, \naas leverages the biologically-inspired evolution-based algorithm rather than meta-controller-based algorithm to improve the sample efficiency. It keeps improving the quality of the candidate population by ruling out the inferior and generating from the fittest. Thanks to the low search cost, \naas can be easily integrated with hardware-aware NAS algorithm by adding another optimization level (Figure~\ref{fig:naas}), achieving the joint search.

Extensive experiments verify the effectiveness of our framework. Under the same \#PE and on-chip memory constraints, the \naas is able to deliver 2.6\X, 4.4\X speedup and 2.1\X, 1.4\X energy savings on average compared to Eyeriss~\cite{chen2016eyeriss}, NVDLA~\cite{nvidia2017dla} design respectively. Integrated with Once-For-All NAS algorithm~\cite{cai2020once}, \naas further improves the top-1 accuracy on ImageNet by 2.7\% without hurting the hardware performance. Using the similar compute resources, \naas achieves 3.0\X, 1.9\X EDP improvements compared to Neural-Hardware Architecture Search~\cite{lin2019nhas}, and NASAIC~\cite{yang2020co} respectively.

\section{\NAAS}
\begin{figure}[t]
    \centering
    \includegraphics[width=\linewidth]{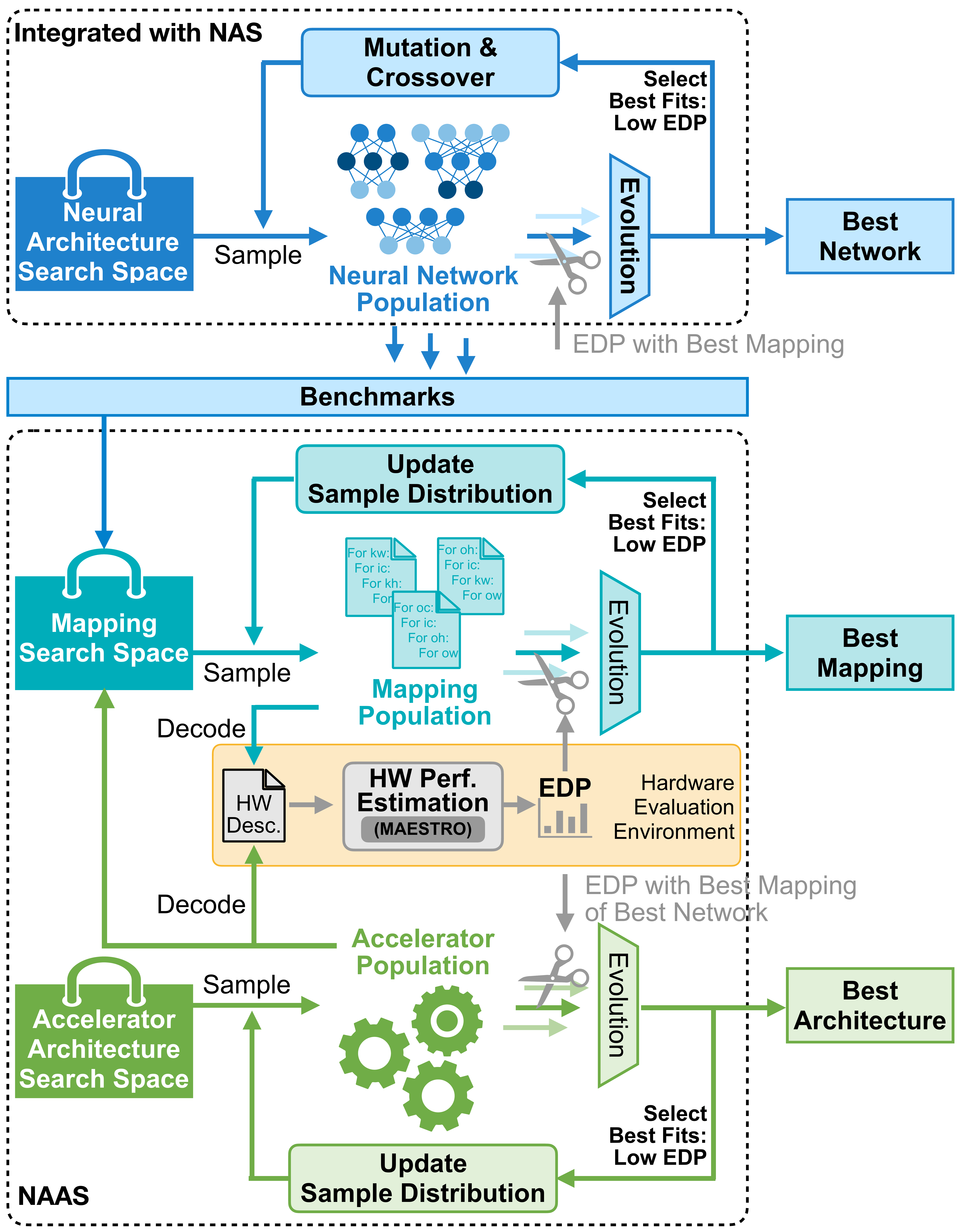}
    \caption{\NAAS.}
    \label{fig:naas}
\end{figure}
Figure~\ref{fig:naas} shows the optimization flow of \NAAS (\naas). \naas explores the design space of accelerators, and compiler's mappings simultaneously.

\subsection{Accelerator Architecture Search}
\label{sec:has}

\begin{figure*}[t]
    \centering
    \includegraphics[width=\linewidth]{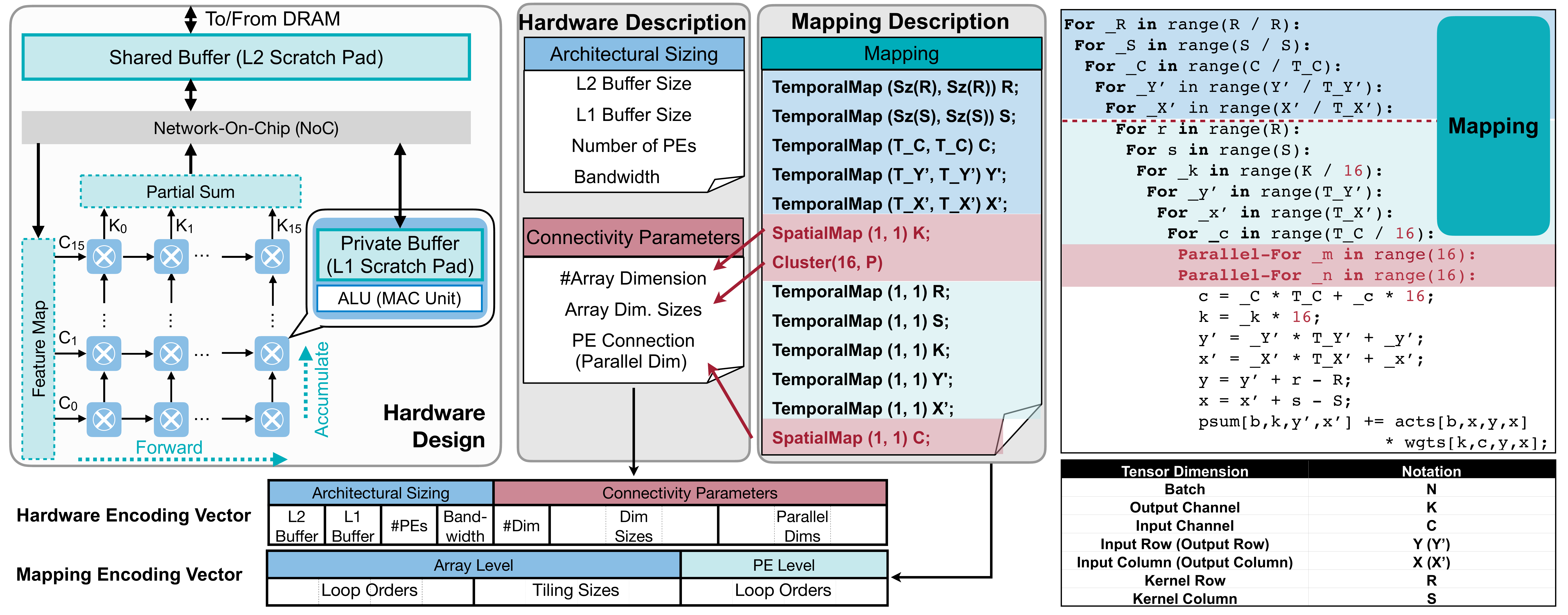}
    \caption{Encoding the accelerator design and compiler mapping into vectors. An accelerator design is described by architectural sizing and connectivity parameters, where PE inter-connection is presented by parallel dimension. A mapping strategy is represented by loop order and corresponding tiling at each dimension of array (the innermost level is PE level). The mapping description in the figure is in  MAESTRO\cite{albericio2016cnvlutin} format which fuses the array parameters and mapping strategy. }
    \label{fig:modeling}
\end{figure*}
\begin{figure}[t]
    \centering
    \includegraphics[width=\linewidth]{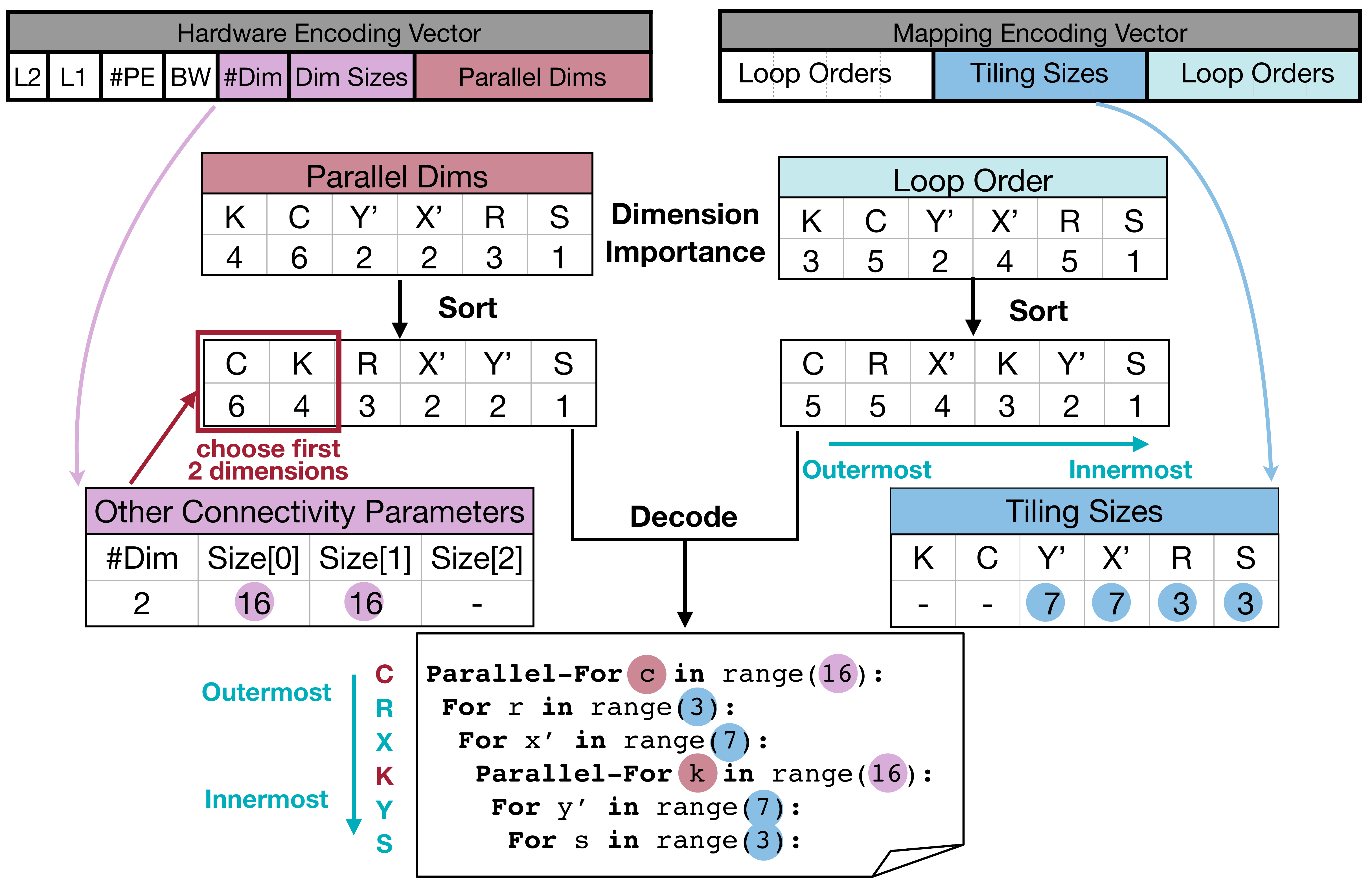}
    \caption{We convert the non-numerical ordering optimization into numerical optimization for parallel dimensions and loop order searching. Both accelerator optimizer and mapping optimizer will generate an importance value for each dimension. The left part shows the parallel dimensions of this 2D array candidate are the dimensions with the largest two importance value. The right part shows the order of each dimension in for-loop is in the decreasing order of its importance value.}
    \label{fig:ordering}
\end{figure}
\begin{figure}[t]
\centering
    \includegraphics[width=0.75\linewidth]{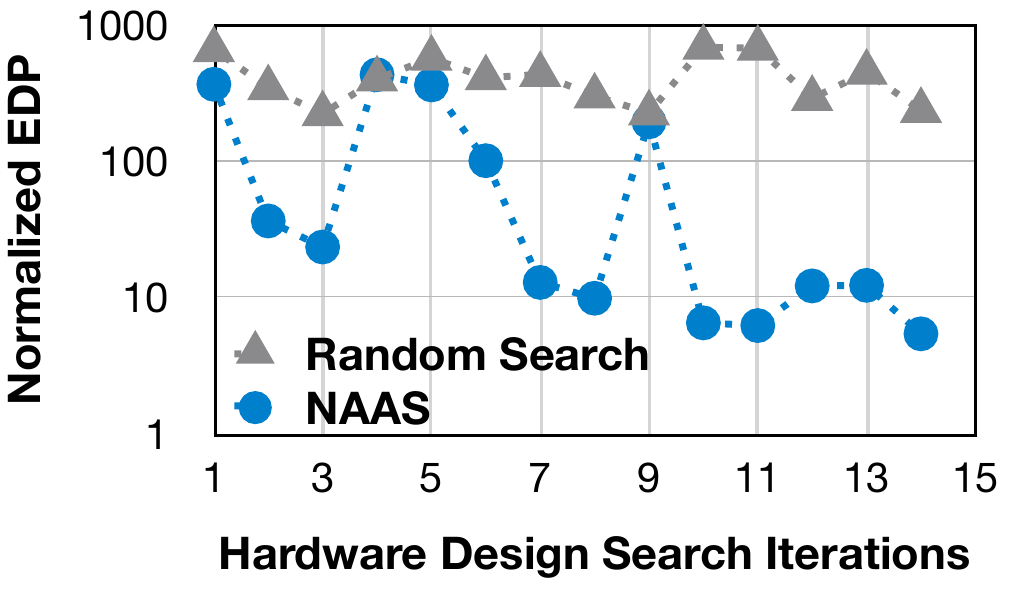}
    \caption{The average of EDP decreases as \naas is learning.}
    \label{fig:has-bitfusion-iter-3d5}
\end{figure}

\paragraph{\textbf{Design Space}} 
The accelerator design knobs can be categorized into two classes:
\begin{enumerate}
    \item Architectural Sizing: the number of processing elements (\#PEs), private scratch pad size (L1 size), global buffer size (L2 size), and memory bandwidth.
    \item Connectivity Parameters: the number of array dimensions (1D, 2D or 3D array), array size at each dimension, and the inter-PE connections.
\end{enumerate}

Most state-of-art searching frameworks only contains architectural sizing parameters in their design space. These sizing parameters are numerical and can be easily embedded into vectors during search. On the other hand,  PE connectivity is difficult to encode as vectors since they are not numerical numbers. Moreover, changing the connectivity requires re-designing the compiler mapping strategies, which extremely increase the searching cost.
In \naas, besides the architectural sizing parameters which are common in other frameworks, we introduce the connectivity parameters into our search space, making it possible to search among 1D, 2D and 3D array as well, and thus our design space includes almost the entire accelerator design space for neural network accelerators.

\paragraph{\textbf{Encoding}}
\label{sec:encoding}
We first model the PE connectivity as the choices of parallel dimensions. For example, parallelism in input channels (C) means a reduction connection of the partial sum register inside each PE. Parallelism in output channels means a broadcast to input feature register inside each PE.
The most straight-forward method to encode the parallel dimension choice is to
enumerate all possible parallelism situations and choose the index of the enumeration as the encoding value. However, since the increment or decrement of indexes does not convey any physical information, it is hard to be optimized. 

To solve this problem, we proposed the ``importance-based'' encoding method for choosing parallelism dimensions in the dataflow and convert the indexing optimization into the sizing optimization. For each dimension, our optimizer will generate an importance value. To get the corresponding parallel dimensions, we first collect all the importance value, then sort them in decreasing order, and select the first \textit{k} dimensions as the parallel dimensions of a \textit{k}-D compute array. 
As shown in the left of Figure~\ref{fig:ordering}, the generated candidate is a 2D array with size $16\times16$. To find the parallel dimension for this 2D array candidate, The importance values are first generated for 6 dimensions in the same way as other numerical parameters in the encoding vector. We then sort the value in decreasing order and determine the new order of the dimensions. Since the importance value of ``C'' and ``K'' are the largest two value, we finally select ``C'' and ``K'' as the parallel dimensions of this 2D array.
The importance value of the dimension represents the priority of the parallelism: a larger value indicates a higher priority and a higher possibility to be paralleled in the computation loop nest, which contains higher relativity with accelerator design compared to indexes of enumerations.

For other numerical parameters, we use the straight-forward encoding method. The whole hardware encoding vector is shown in Figure~\ref{fig:modeling}, which contains all of the necessary parameters to represent an accelerator design paradigm.

\paragraph{\textbf{Evolution Search}}
\label{sec:evo}
We leverage the evolution strategy~\cite{hansen2006cma} to find the best solution during the exploration. In order to take both latency and energy into consideration, we choose the widely used metric Energy-Delay Product (EDP) to evaluate a given accelerator configuration on a specific neural network workload.
At each evolution iteration, we first sample a set of candidates according to a multivariate normal distribution in $\left [ 0, 1 \right ]^{\left | \theta \right |}$. Each candidate is represented as a $\left | \theta \right |$-dimension vector. These candidates are then projected to hardware encoding vectors and decoded into accelerator design. We rule out the invalid accelerator samples and keep sampling until the candidate set reaches a predefined size (population size) in our experiments. 
To evaluate the candidate performance, we need to perform mapping strategy search in Section~\ref{sec:mss} on each benchmark and adopt the best searched result as the EDP reward of this candidate. After evaluating all the candidates on the benchmarks, we update the sampling distribution based on the relative ordering of their EDP. Specifically, we select the top solutions as the ``parents'' of the next generation and use their center to generate the new mean of the sampling distribution. We update the covariance matrix of the distribution to increase the likelihood of generating samples near the parents~\cite{hansen2006cma}. We then repeat such sampling and updating process. 

Figure~\ref{fig:has-bitfusion-iter-3d5} shows the statistics of energy-delay products of hardware candidates in the population. As the optimization continues,  the EDP mean of \naas candidates decreases while that of random search remains high, which indicates that \naas gradually improves the range of hardware selections.
\subsection{Compiler Mapping Strategy Search}
\label{sec:mss}
The performance and energy efficiency of deep learning accelerators also depend on how to \textit{map} the neural network task on the accelerator. The search space of compiler mapping strategy is much larger than accelerator design, since different convolution layers may not share the same optimal mapping strategy. Hence we optimize the mapping for each layer independently using the similar evolution-based search algorithm to accelerator design search in Section~\ref{sec:evo}.

The compiler mapping strategy consists of two components: the execution order and the tiling size of each for-loop dimension.
Similar to the accelerator design search, the order of for-loop dimensions is non-trivial. Rather than enumerating all of the possible execution orders and using indexes as encoding, we use the similar ``importance-based'' encoding methods in Section~\ref{sec:encoding}. For each dimension of the array (corresponding to each level of for-loop nests), the mapping optimizer will assign each convolution dimension with an importance value. The dimension with the highest importance will become the outermost loop while the one with the lowest importance will be placed at the innermost in the loop nests. 
The right of Figure~\ref{fig:ordering} gives an example. The optimizer firstly generates the importance values for 6 dimensions, then sort the value in decreasing order and determine the corresponding order of the dimensions. Since ``C'' and ``R'' dimension have largest value 5, they will become the outermost loops. ``S'' dimension has the smallest value 1, so it is the innermost dimension in the loops. 
This strategy is interpretable, since the importance value represents the data locality of the dimension: the dimension labeled as most important has the best data locality since it is the outermost loop, while the dimension labeled as least important has the poorest data locality therefore it is the innermost loop.

As for tiling sizes, since they are highly related to the network parameters, we use the scaling ratio rather than the absolute tiling value. Hence, the tiling sizes are still numerical parameters and able to adapt to different networks. The right part of figure~\ref{fig:ordering} illustrates the composition of the mapping encoding vector. Note that for PE level we need to ensure that there is only one MAC in a PE, so we only search the loop order at PE level. For each array level, the encoding vector contains both the execution order and tiling size for each for-loop dimension.

\subsection{Integrated with Neural Architecture Search}
\label{sec:qnas}

Thanks to the low search cost, we can integrate our framework with neural architecture search to achieve neural-accelerator-compiler co-design. Figure~\ref{fig:naas} illustrates integrating \naas with NAS. The joint design space is huge, and in order to improve the search efficiency, we choose and adapt Once-For-All NAS algorithm for \naas. First, \naas generates a pool of accelerator candidates. For each accelerator candidate, we sample a network candidate from NAS framework which satisfies the pre-defined accuracy requirement. Since each subnet of Once-For-All network is well trained, the accuracy evaluation is fast. We then apply the compiler mapping strategy search for the network candidate on the corresponding accelerator candidate. NAS optimizer will update using the searched EDP as a reward. Until NAS optimizer reaches its iteration limitations, and feedback the EDP of the best network candidate to accelerator design optimizer. We repeated the process until the best-fitted design is found. In the end, we obtain a tuple of matched accelerator, neural network, and its mapping strategy with guaranteed accuracy and lowest EDP.
\section{Evaluation}
We evaluate \NAAS's performance improvement step by step: 1) the improvement from applying \naas  given the same hardware resource; 2) performance of \naas which integrates the Once-For-All to achieve the neural-accelerator-compiler co-design.

\begin{figure*}[t]
\centering
    \includegraphics[width=0.9\linewidth]{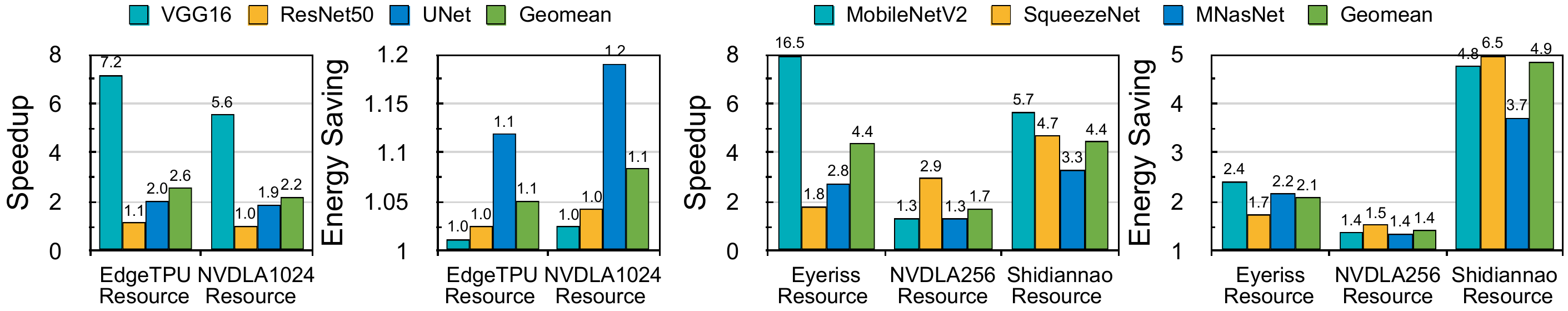}
    \caption{Speedup and energy savings when \naas searches the accelerator and mapping for a set of network benchmarks within the baseline hardware resources. For instance, using the same \#PEs and SRAM size as EdgeTPU, the accelerator designed by \naas runs VGG16, ResNet50 and UNet 2.6$\times$ faster than EdgeTPU on average.}
    \label{fig:has-geo}
\end{figure*}
\begin{figure*}[t]
\centering
    \includegraphics[width=0.9\linewidth]{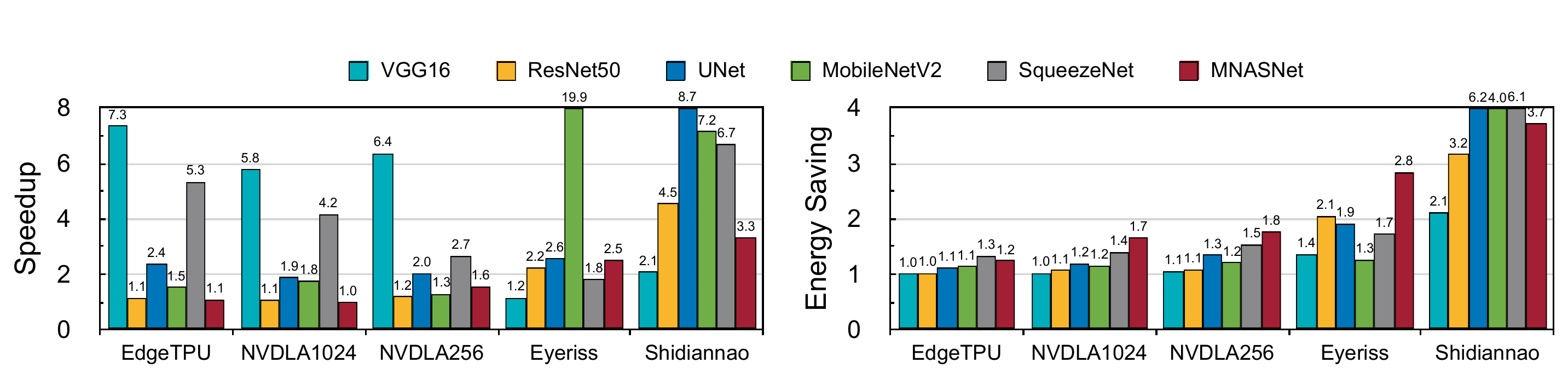}
    \caption{Speedup and energy savings when \naas searches the accelerator and mapping for single network benchmark within the baseline hardware resources.}
    \label{fig:has-one}
\end{figure*}
\begin{figure*}[t]
\centering
    \includegraphics[width=0.9\linewidth]{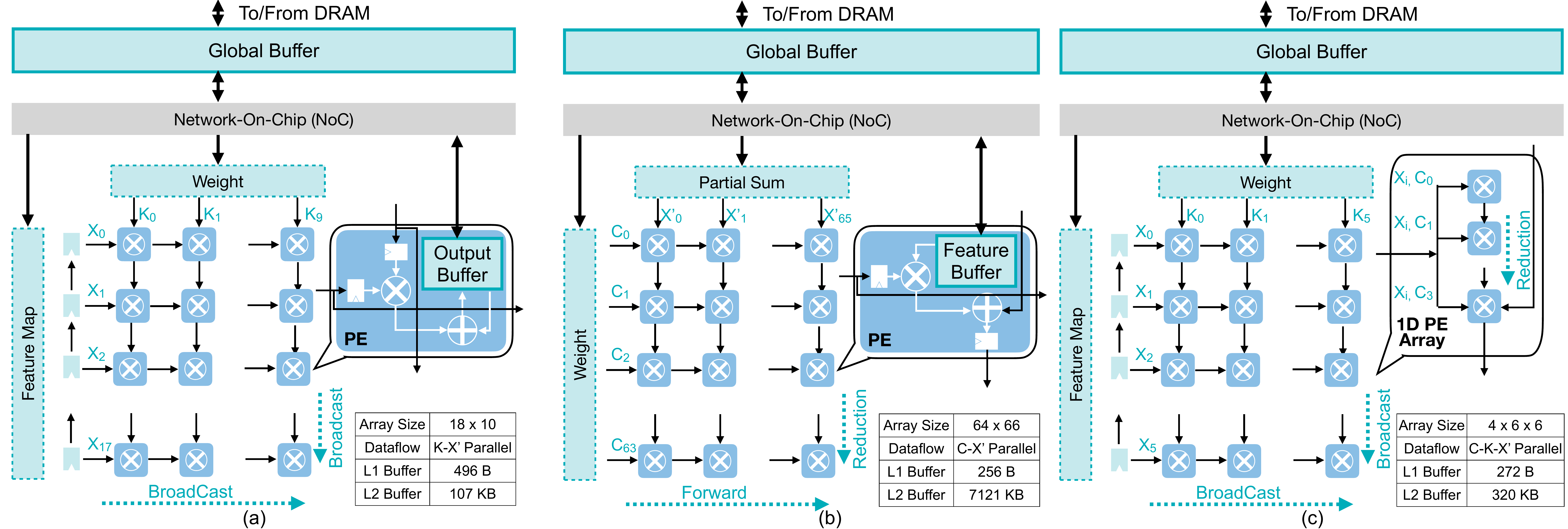}
    \caption{\naas offers different architectures for different networks when given different computation resources: (a) 2D array with K-X' parallelism for ResNet using Eyeriss resource; (b) 2D array with C-X' parallelism for VGG16 using EdgeTPU resource; (c) 3D array with C-K-X' parallelism for VGG16 using Shidiannao resource.}
    \label{fig:has-archs}
\end{figure*}
\begin{figure}[t]
\centering
    \includegraphics[width=0.75\linewidth]{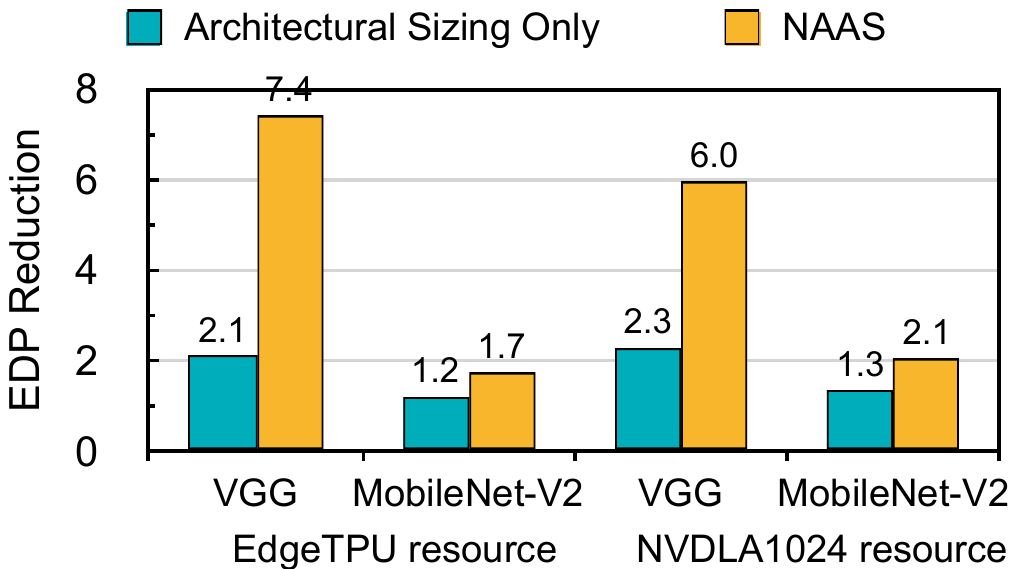}
    \caption{Compared to searching the architectural sizing only~\cite{yang2020co, lin2019nhas}, searching the connectivity parameters and mapping strategies as well achieves considerable EDP reduction.}
    \label{fig:no-ordering}
\end{figure}
\begin{figure}[t]
\centering
    \includegraphics[width=0.8\linewidth]{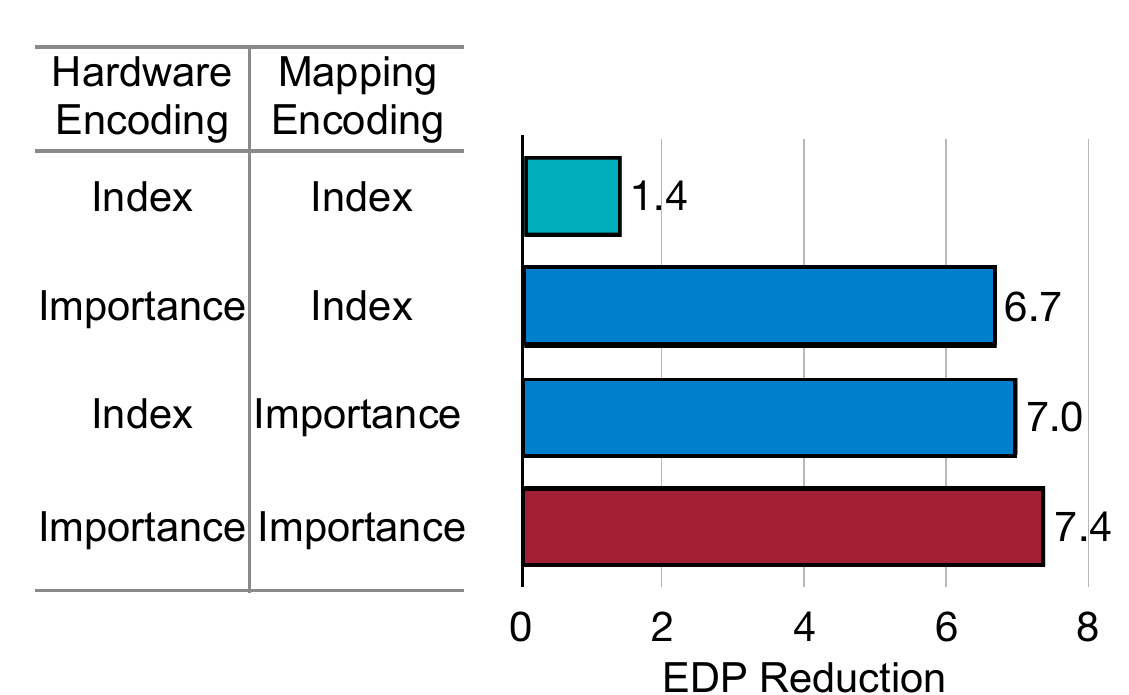}
    \caption{Compared to index-based encoding, importance-based encoding achieves better EDP reduction.}
    \label{fig:importance}
\end{figure}

\begin{figure}[t]
\centering
    \includegraphics[width=0.5\linewidth]{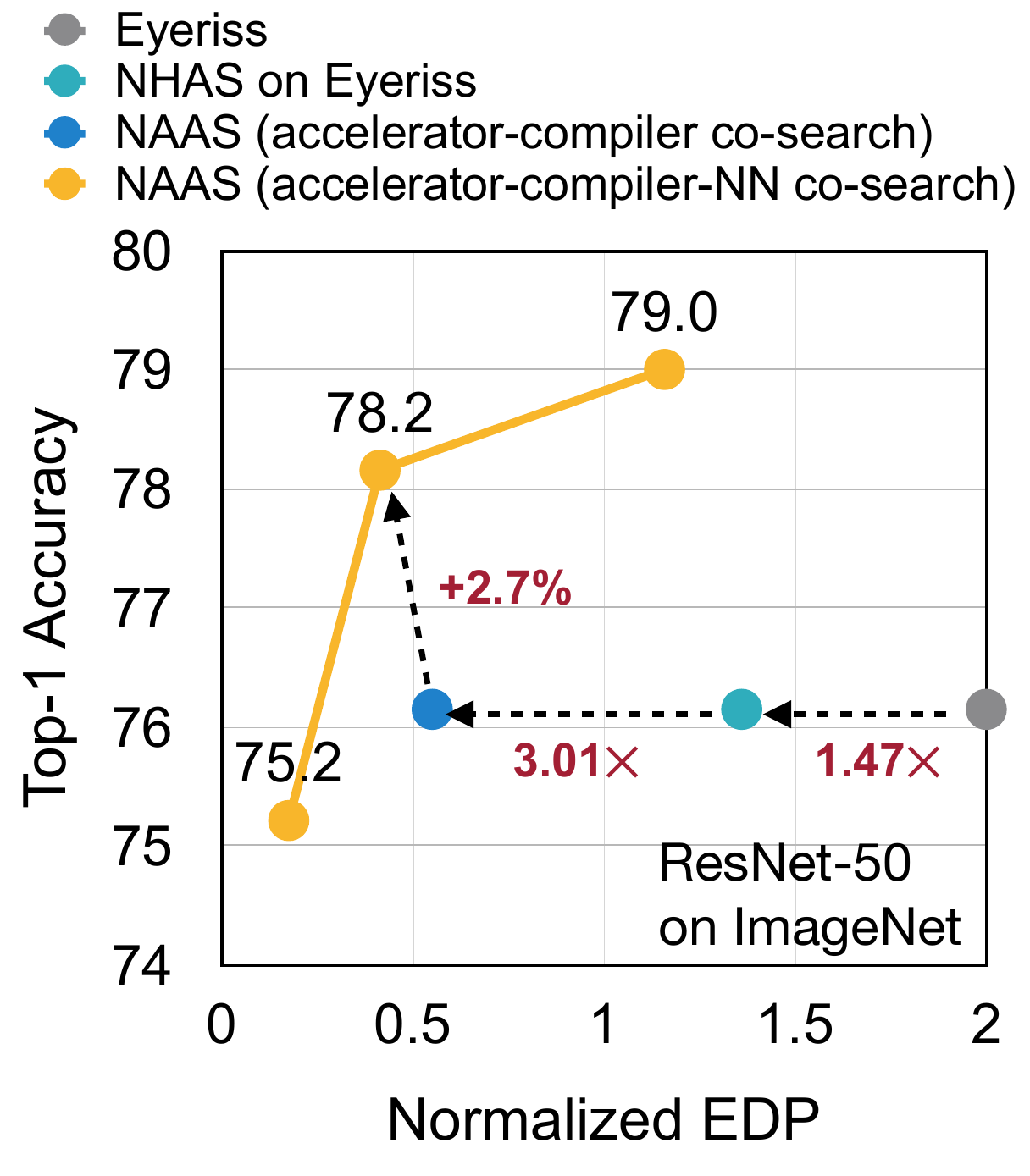}
    \caption{Accuracy \vs Normalized EDP on ImageNet (batch = 1) under Eyeriss hardware resource. Integrating NAAS with OFA NAS further improves model accuracy.}
    \label{fig:qnas-res}
\end{figure}

\subsection{Evaluation Environment}
\label{sec:eval-env}
\paragraph{\textbf{Design Space of \naas}}
We select four different resource constraints based on EdgeTPU, NVDLA~\cite{nvidia2017dla}, Eyeriss~\cite{chen2016eyeriss} and Shidiannao~\cite{du2015shidiannao}. When comparing to each baseline architecture, \naas is conducted within corresponding computation resource constraint including the maximum \#PEs, the maximum total on-chip memory size, and the NoC bandwidth. \naas searches \#PEs at stride of 8, buffer sizes at stride of 16B, array sizes at stride of 2. 

\paragraph{\textbf{CNN Benchmarks}} 
We select 6 widely-used CNN models as our benchmarks. The benchmarks are divided into two sets: classic large-scale networks (VGG16, ResNet50, UNet) and light-weight efficient mobile networks (MobileNetV2, SqueezeNet, MNasNet). Five deployment scenarios are divided accordingly: we conduct \naas for large models with more hardware resources (EdgeTPU, NVDLA with 1024 PEs), and for small models  with limited hardware resources (ShiDianNao, Eyeriss, and NVDLA with 256 PEs).

\paragraph{\textbf{Design Space in Once-For-All NAS}}
When integrating with Once-For-All NAS, the neural architecture space is modified from ResNet-50 design space following the open-sourced library~\cite{cai2020once}. There are 3 width multiplier choices (0.65, 0.8, 1.0) and 18 residual blocks at maximum, where each block consists of three convolutions and has 3 reduction ratios (0.2, 0.25, 0.35). Input image size ranges from 128 to 256 at stride of 16. In total there are $10^{13}$ possible neural architectures.

\subsection{Improvement from \naas}

Figure~\ref{fig:has-geo} shows the speedup and energy savings of \naas using the same hardware resources compared to the baseline architectures. When running large-scale models, \naas delivers 2.6\X, 2.2\X speedup and 1.1\X, 1.1\X energy savings on average compared to EdgeTPU and NVDLA-1024 . Though \naas tries to provide a balanced performance on all benchmarks by using geomean EDP as reward, VGG16 workload sees the highest gains from \naas. When inferencing light-weight models, \naas achieves 4.4\X, 1.7\X, 4.4\X speedup and 2.1\X, 1.4\X, 4.9\X energy savings on average compared to Eyeriss, NVDLA-256, and ShiDianNao. Similar to searching for large models, different models obtain different benefits from \naas under different resource constraints.

Figure~\ref{fig:has-archs} demonstrates three examples of searched architectures. When given different computation resources, for different NN models, \naas provides different solutions beyond numerical design parameter tuning. Different dataflow parallelisms determine the different PE micro-architecture and thus PE connectivies and even feature/weight/partial-sum buffer placement. 

Figure~\ref{fig:no-ordering} illustrates the benefits of searching connectivity parameters and mapping strategy compared to searching architectural sizing only. \naas outperforms architectural sizing search by 3.52\x, 1.42\x EDP reduction on VGG and MobileNetV2 within EdgeTPU resources, as well as 2.61\x, 1.62\x improvement under NVDLA-1024 resources.  

Figure~\ref{fig:importance} further shows the EDP reduction using different encoding methods for non-numerical parameters in hardware and mapping encoding vectors. Our proposed importance-based encoding method significantly improves the performance of optimization by reducing EDP from 1.4\x to 7.4\x.

\subsection{More Improvement from Integrating NAS }

Different from accelerator architectures, neural architectures have much more knobs to tune (\eg, network depths, channel numbers, input image sizes), providing us with more room to optimize. 
To illustrate the benefit of \naas with NAS, we evaluate on ResNet50 with hardware resources similar to Eyeriss. Figure~\ref{fig:qnas-res} shows that \naas (accelerator only) outperforms Neural-Hardware Architecture Search (NHAS)~\cite{lin2019nhas} (which only searches the neural architecture and the accelerator architectural sizing) by 3.01\X EDP improvement. By integrating with neural architecture search, \naas achieves 4.88\X EDP improvement in total as well as 2.7\% top-1 accuracy improvement on ImageNet dataset than Eyeriss running ResNet50. 

\paragraph{\textbf{Comparison to NASAIC}} We also compare our \naas with previous work NASAIC~\cite{yang2020co} in Table~\ref{tab:nasaic}. NASAIC adopts DLA~\cite{nvidia2017dla}, ShiDianNao~\cite{du2015shidiannao} as source architectures for its heterogeneous accelerator, and only searches the allocation of \#PEs and NoC bandwidth. In contrast, we explore more possible improvements by adapting both accelerator design searching and mapping strategy searching. Inferencing the same network searched by NASAIC, \naas outperforms NASAIC by 1.88\X EDP improvement (3.75\X latency improvement with double energy cost) using the same design constraints.

\begin{table}[t]
  \renewcommand{\arraystretch}{1.15}
\caption{\naas (accelerator only) achieves better EDP.}
\scalebox{0.75}{
\begin{tabular}{c|c|cccc}
\toprule
\textbf{Search} & \multirow{2}{*}{\textbf{Arch}}   & \textbf{Cifar-10} & \textbf{Latency}  & \textbf{Energy} & \textbf{EDP}\\
 \textbf{Approach} &  & \textbf{Accuracy} & \textbf{(cycles)} & \textbf{(nJ)} & \textbf{(cycles$\cdot$nJ)} \\ \midrule
\multirow{2}{*}{NASAIC~\cite{yang2020co}} & DLA~\cite{nvidia2017dla}  & 93.2 & \multirow{2}{*}{3e5} & \multirow{2}{*}{1e9} & \multirow{2}{*}{3e14}   \\
  & Shi~\cite{du2015shidiannao}  &  91.1 & & \\ \hline
\naas & DLA~\cite{nvidia2017dla}  & 93.2 & 8e4 & 2e9 & 2e14 \\ \bottomrule
\end{tabular}
}
\label{tab:nasaic}
\end{table}
\begin{table}[t]
\caption{We achieve much lower search cost on ImageNet (Gds: GPU days. \textit{N}: number of development scenarios.)}
  \renewcommand*{\arraystretch}{1.15}
\scalebox{0.65}{
\begin{tabular}{c|ccc|cc}
\toprule
\multirow{2}{*}{Approach} & Co-Search &  NN Training & Total & AWS & CO$_2$\\
  & Cost (Gds) & Cost (Gds) & Cost (Gds) & Cost & Emission \\
\midrule
NASAIC 
& 500\x12\textit{N}~=~6000\textit{N} & 16\textit{N} & 6000\textit{N} & \$441,000\textit{N} & 41,000\textit{N} lbs \\
NHAS & 12~+~4\textit{N} & 16\textit{N} & 12~+~20\textit{N} & \$1,500\textit{N} & 150\textit{N} lbs \\
Ours & $<$~0.25\textit{N} & 50 & $<$~50~+~0.25\textit{N} & $<$~\$18\textit{N} & $<$~2\textit{N} lbs \\
\bottomrule
\multicolumn{6}{l}{* NASAIC's search cost is an optimistic projection from Cifar10.} \\
\multicolumn{6}{l}{* AWS cost \$75/Gd and CO$_2$ Emission is 7.5 lbs/Gd.}
\end{tabular}
}
\label{tab:search-cost}
\end{table}

\paragraph{\textbf{Search Cost}} 
Table~\ref{tab:search-cost} reports the search cost of our \naas compared to NASAIC and NHAS when developing accelerator and network for $N$ development scenarios, where ``AWS Cost" is calculated based on the price of on-demand P3.16xlarge instances, and ``CO$_2$ Emission" is calculated based on Strubell \etal ~\cite{strubell2019energy}. 
NASAIC relies on a meta-controller-based search algorithm which requires training every neural architecture candidates from scratch. 
NHAS~\cite{lin2019nhas} decouples training and searching in NAS but it also requires retraining the searched network for each deployment which costs 16$N$ GPU days. The high sample efficiency of \naas makes it possible to integrate the Once-For-All NAS in one search loop. As a conservative estimation, \naas saves more than 120\X search cost compared to NASAIC on ImageNet.
\section{Related Works}

\paragraph{\textbf{Accelerator Design-Space Exploration}} Ealier work  focuses on fine-grained hardware resource assignment for deployment on FPGAs~\cite{hao2019fpga, motamedi2016design, zhong2017design, chen2019cloud, kao2020confuciux, jiang2020hardware}. 
Several work focuses on co-designing neural architectures and ASIC accelerators. 
Yang~\etal~\cite{yang2020co} (NASAIC) devise a controller that simultaneously predicts neural architectures as well as the selection policy of various IPs in a heterogeneous accelerator. Lin~\etal~\cite{lin2019nhas} focuses on optimizing the micro architecture parameters such as the array size and buffer size for given accelerator design while searching the quantized neural architecture. Besides, some work focuses on optimizing compiler mapping strategy on fixed architectures. Chen~\etal~\cite{chen2018tvm} (TVM) designed a searching framework for optimizing for-loop execution on CPU/GPU or other fixed platforms. Mu~\etal~\cite{mu2020history} proposed a new searching algorithm for tuning mapping strategy on fixed GPU architecture.
On the contrary, our work explores not only the sizing parameters but also the connectivity parameters and the compiler mapping strategy. We also explore the neural architecture space to further improve the performance.
Plenty of work provides the modeling platform for design space exploration \cite{shao2015aladdin, wu2019accelergy, parashar2019timeloop,  kwon2019understanding}.
We choose MAESTRO~\cite{kwon2019understanding} as the accelerator evaluation backend.

\paragraph{\textbf{AutoML and Hardware-Aware NAS}}

Researchers have looked to automate the neural network design using AutoML.  Grid search~\cite{kwon2019understanding, tan2019efficientnet} and  reinforcement learning with meta-controller~\cite{zoph2016neural, lu2019neural, yang2020co} both suffer from prohibitive search cost. One-shot-network-based frameworks~\cite{cai2018proxylessnas, wang2019haq, guo2019single} achieved high performance at a relatively low search cost. These NAS algorithms require retraining the searched networks while Cai~\etal~\cite{cai2020once} proposed Once-For-All network of which the subnets are well trained and can be directly extracted for deployment. Recent neural architecture search (NAS) frameworks started to incorporate the hardware into the search feedback loop~\cite{lu2019neural, zhang2019skynet, cai2018proxylessnas, li2020edd, wang2019haq, cai2020once}, though they have not explored hardware accelerator optimization.

\section{Conclusion}
We propose an evolution-based accelerator-compiler co-search framework, \naas. It not only searches the architecture sizing parameters but also the PE connectivity and compiler mapping strategy. Integrated with the  Once-for-All NAS algorithm, it explores the search spaces of neural architectures, accelerator architectures, and mapping strategies together while reducing the search cost by 120\X compared with previous work.
Extensive experiments verify the effectiveness of \naas. Within the same computation resources as Eyeriss~\cite{chen2016eyeriss}, \naas provides 4.4\X energy-delay-product reduction with 2.7\% top-1 accuracy improvement on ImageNet dataset compared to directly running ResNet-50 on Eyeriss. Using the similar computation resources, \naas integrated with NAS achieves 3.0\X, 1.9\X EDP improvements compared to NHAS~\cite{lin2019nhas}, and NASAIC~\cite{yang2020co} respectively.

\noindent\textbf{Acknowledgements.}
This work was supported by NSF CAREER Award \#1943349 and SRC GRC program under task 2944.001. We also thank AWS Machine Learning Research Awards for the computational resource.




\bibliography{reference}
\bibliographystyle{IEEEtran}

\end{document}